\documentclass[runningheads]{llncs}

\usepackage[mobile]{eccv}

\usepackage{eccvabbrv}

\usepackage{graphicx}
\usepackage{booktabs}

\usepackage{hyperref}

\begin{document}

\title{TexGen: Text-Guided 3D Texture Generation with Multi-view Sampling and Resampling} 

\titlerunning{TexGen}

\author{Dong Huo\inst{1,3*} \and
Zixin Guo\inst{2} \and
Xinxin Zuo\inst{3}\and
Zhihao Shi\inst{3} \and
Juwei Lu\inst{3} \and
Peng Dai\inst{3} \and
Songcen Xu\inst{3}\and
Li Cheng\inst{1}\and
Yee-Hong Yang\inst{1}
}

\authorrunning{D.~Huo et al.}

\institute{$^{1}$ University of Alberta, Canada\\
\email{\{dhuo, lcheng5\}@ualberta.ca, yang@cs.ualberta.ca}\\
$^{2}$ University of Toronto, Canada\\
\email{zixin.guo@mail.utoronto.ca}\\
$^{3}$ Huawei Noah's Ark Lab\\
\email{\{xinxin.zuo1, zhihao.shi, juwei.lu, peng.dai, xusongcen\}@huawei.com}
}

\onecolumn{
 \maketitle
 \centerline{
 \includegraphics[width=\linewidth]{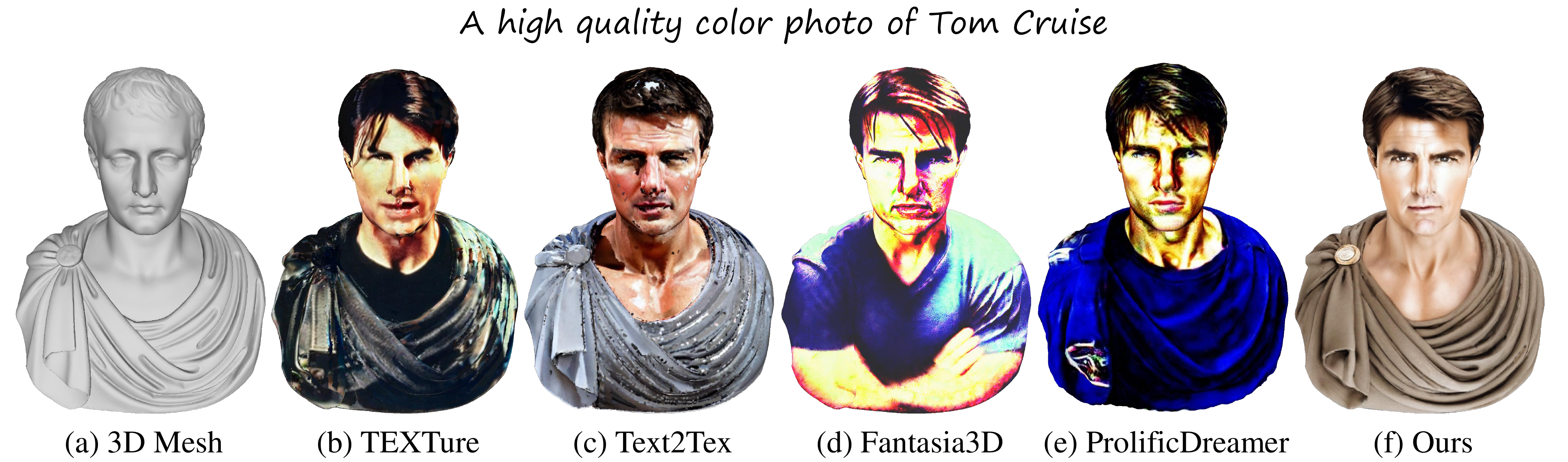}
}
\captionof{figure}{Given a 3D mesh, we present text-driven texture generation results from previous state-of-the-art approaches (TEXTure~\cite{richardson2023texture}, Text2Tex~\cite{chen2023text2tex}, Fantasia3D~\cite{chen2023fantasia3d}, and ProlificDreamer~\cite{wang2023prolificdreamer}) as well as our proposed method.}
\label{fig:teaser}
}
\let\thefootnote\relax\footnotetext{\noindent * Work done during an internship at Huawei Noah's Ark Lab}

\begin{abstract}

Given a 3D mesh, we aim to synthesize 3D textures that correspond to arbitrary textual descriptions. Current methods for generating and assembling textures from sampled views often result in prominent seams or excessive smoothing. To tackle these issues, we present TexGen, a novel multi-view sampling and resampling framework for texture generation leveraging a pre-trained text-to-image diffusion model. For view consistent sampling, first of all we maintain a texture map in RGB space that is parameterized by the denoising step and updated after each sampling step of the diffusion model to progressively reduce the view discrepancy. An attention-guided multi-view sampling strategy is exploited to broadcast the appearance information across views. To preserve texture details, we develop a noise resampling technique that aids in the estimation of noise, generating inputs for subsequent denoising steps, as directed by the text prompt and current texture map.
Through an extensive amount of qualitative and quantitative evaluations, we demonstrate that our proposed method produces significantly better texture quality for diverse 3D objects with a high degree of view consistency and rich appearance details, outperforming current state-of-the-art methods. 
Furthermore, our proposed texture generation technique can also be applied to texture editing while preserving the original identity. More experimental results are available at \href{https://dong-huo.github.io/TexGen/}{https://dong-huo.github.io/TexGen/}.

\end{abstract}    
\section{Introduction}
\label{sec:intro}
Generating high-quality 3D content is an essential component of visual applications in films, games, and upcoming AR/VR industries. While many prior works on 3D synthesis have focused on the geometric components of the assets, textures have garnered less attention which play a vital role in enhancing the realism of 3D assets. In this paper, we aim to realize automatic text-driven 3D texture synthesis for various meshes.

Recently, the research community has witnessed remarkable progress in text-to-image (T2I) generation ~\cite{ho2020denoising, song2020denoising, ho2022classifier, rombach2022high}. However, the generation of 3D assets still faces challenges due to the limited size of 3D datasets~\cite{deitke2023objaverse, wu2023omniobject3d, chang2015shapenet}, characterized by overly simplified textures.
To this end, existing methods have been leveraging the visual information encoded in the 2D image priors of pre-trained T2I diffusion models. A thread of studies, such as score distillation sampling (SDS) and variational score distillation (VSD)~\cite{poole2022dreamfusion, lin2023magic3d, chen2023fantasia3d,wang2023prolificdreamer}, aim to distil the diffusion priors as score functions to directly optimize a 3D representation (\eg, geometry and texture), ensuring that its rendered outputs align well with the high-likelihood image priors. Despite remarkable successes in 2D-to-3D conversion, there are noticeable shortcomings. Specifically, textures generated using score distillation pipelines tend to be over-saturated, as illustrated in Fig.\ref{fig:teaser}(d), or suffer from issues such as blurry edges and color artifacts, as seen in Fig.\ref{fig:teaser}(e).

Beyond score distillation methods, another thread of studies for texture synthesis involves mapping multi-view images, generated from the T2I models, onto a global UV texture. For example, TEXTure\cite{richardson2023texture} and Text2tex\cite{chen2023text2tex} adopted an autoregressive image inpainting pipeline to progressively assemble multi-view images generated from T2I models. 
While these methods can produce high-fidelity textures for particular views, they often exhibit noticeable seams on the assembled texture map, as evident in Fig.\ref{fig:teaser}(b) and (c). This issue arises due to error accumulation stemming from the autoregressive view inpainting process, and the primary cause of this error lies in the limited guidance provided by previously observed views~\cite{cao2023texfusion}. 
Considering the sequential characteristics of the denoising process, TexFusion~\cite{cao2023texfusion} proposed to mitigate the inconsistency of different views at each sampling step of the diffusion model. They conducted a interlaced multiview sampling in latent space, after which they decoded the latent map to RGB image for each view independently via the pre-trained stable diffusion VAE decoder. However, the view inconsistencies recurred after the view independent decoding. A shallow MLP was optimized in the RGB texture space to smooth out the inconsistency, which will result in over-smoothed textures. 


In response to the challenges of view inconsistency and over-smoothness in texture generation, we introduce \textbf{TexGen}, a novel multi-view sampling and resampling strategy that directly generates view consistent RGB images with rich appearance details from pre-trained 2D T2I models for texture assembling. In detail, we derive a UV texture map in the RGB space that will be iteratively updated during each denoising step, to gradually unveil texture details. At each denoising step, we predict the latent \textit{denoised observations} of sequentially sampled views around the 3D objects. These denoised observations are then decoded and assembled onto the texture map, enabling direct generation of view-consistent RGB textures without the need for additional MLP optimization steps~\cite{cao2023texfusion}. An attention-guided multi-view sampling mechanism is proposed to ensure better appearance consistency across views within each denoising step. 

More importantly, we develop a Text\&Texture-Guided Resampling approach for noise estimation which leverages the information from both the renders of view consistent texture map at each denoising step and the high-frequency priors from the pre-trained T2I model.  
Through the fusion of texture and text-guided noise estimation, our generated textures not only maintain view consistency but also exhibit a rich diversity of details.

In summary, our key contributions can be outlined as follows: (1) we propose a multi-view sampling and resampling framework for text-driven texture generation using pre-trained 2D T2I models;
(2) in particular, our proposed attention-guided multi-view sampling as well as text\&texture-guided noise resampling techniques ensure that both 3D view consistency and rich details are preserved in the generated textures; 
(3) we demonstrate the effectiveness of our approach in texturing diverse 3D objects, showcasing superior performance compared to state-of-the-art methods. It is noteworthy that our proposed framework can naturally support text-driven texture editing as well.

\section{Related Work}
\label{sec:review}

\subsection{Diffusion Models in 3D Domain}
  
Inspired by the success of 2D image generation with diffusion models, researchers have also attempted to utilize diffusion models to generate 3D objects in the form of various representations, such as point clouds~\cite{zeng2022lion,luo2021diffusion,zhou20213d, nichol2022point}, and neural fields~\cite{nam20223d,wang2023rodin}. For example, Point$\cdot$E~\cite{nichol2022point} trains a diffusion model using a large synthetic 3D dataset to produce a 3D RGB point cloud conditioned on a synthesized single view from a text prompt. However, these works mainly focus on geometry generation and do not specifically tackle 3D texture synthesis. Yu~\etal~\cite{yu2023texture} trains a diffusion model for mesh texture generation of specific object categories. Although Shap$\cdot$E~\cite{jun2023shap} is proposed to directly generate the parameters of implicit functions that can be rendered as both textured meshes and neural radiance fields, it cannot generalize to incorporate arbitrary text prompts. Moreover, the generated textures tend to be over smoothed with rather low quality as compared with the generated images from the T2I model. 


\subsection{Lifting pre-trained 2D generative models to 3D}  

Initially, the process of distilling 3D objects from pre-trained 2D models has been enhanced by the development of joint text-image embedding, such as Contrastive Language-Image
Pre-training (CLIP)~\cite{radford2021learning}. For example, 
CLIP-Mesh~\cite{mohammad2022clip} learns to generate a mesh with the guidance of the CLIP text embedding and the corresponding image embedding of the diffusion model. However, since the CLIP guidance is rather sparse, the generated 3D models for CLIP-based approaches~\cite{sanghi2023clip, michel2022text2mesh,lei2022tango} are rather coarse. 

Recently, researchers have leveraged large-scale 2D T2I diffusion models to distil individual 3D objects in the form of neural radiance fields. Among various distilling approaches, a dominant one is Score Distillation Sampling (SDS)~\cite{poole2022dreamfusion}. SDS pioneered the approach with many follow-up works~\cite{lin2023magic3d,metzer2023latent,chen2023fantasia3d,wang2023prolificdreamer,tang2023make,sun2023dreamcraft3d,guo2024decorate3d}. 
For example, Magic3D~\cite{lin2023magic3d} proposed a coarse-to-fine strategy to improve generation quality. Latent-NeRF~\cite{metzer2023latent} performed distillation in the latent space of latent diffusion model (LDM)~\cite{rombach2022high}. A crucial drawback of this line of work is that SDS typically requires strong guidance, resulting in low diversity and over-saturation of the generated textures.
ProlificDreamer~\cite{wang2023prolificdreamer} proposed to address this issue with a Variational Score Distillation (VSD) algorithm that adopts a particle-based variational inference to estimate the distribution of 3D scenes instead of a single point as in SDS. Yet, it still suffers from issues like blurry edges and color artefacts.


\noindent\textbf{Texture Synthesis with Multiview Denoising.} 
Instead of relying on the lengthy optimization of score distillation pipelines, an alternative research direction is directly leveraging the sampling process in diffusion models to synthesize UV textures. TEXTure~\cite{richardson2023texture} and Text2tex~\cite{chen2023text2tex} adopt a depth-aware diffusion model~\cite{rombach2022high, zhang2023adding} to progressively paint the mesh surface from different views and aggregate the images generated from the T2I model of sampled views into the texture map. 
While rich textures and details can be faithfully synthesized, there were obvious seams on the assembled texture map due to error accumulation in the process of the autoregressive view update. 
To alleviate this problem, TexFusion~\cite{cao2023texfusion} proposed to interleave texture assembling with denoising steps in different camera views and maintained a latent texture map at each sampling step. To convert latent features to RGB textures, they optimized an intermediate neural color field on the decoding of 2D renders of the latent texture which would wash out the rich details ~\cite{muller2022instant}. Our proposed approach distinguishes itself from previous methods with its ability to generate 3D-consistent textures while preserving rich details in the meantime.




\section{Proposed Method}
\label{sec:method}
\subsection{Overview}
In this section, we present an overview of our proposed multi-view sampling and resampling strategy to synthesize view-consistent textures from a pre-trained T2I diffusion model. We first introduce the sampling process of the Denoising Diffusion Implicit Models (DDIM)~\cite{song2020denoising}, which forms the basis of our texture sampling approach. 

\noindent\textbf{DDIM Sampling.} 
Assuming we sequentially sample $N$ distinct views around a 3D mesh, the DDIM sampling process for each sampled viewpoint $i$ at the denoising step $t$ can be described as follows:
\begin{equation}
x_{t-1}^i = \sqrt{\alpha_{t-1}}\cdot \hat{x}_0^i(x_t^i) + \sqrt{1 - \alpha_{t - 1}}\cdot \epsilon_\theta(x_t^i),
\label{eqn:ddim_1}
\end{equation}
with
\begin{equation}
\hat{x}_0^i(x_t^i) = \frac{x_t^i - \sqrt{1 - \alpha_t}\cdot\epsilon_\theta(x_t^i)}{\sqrt{\alpha_t}},
\label{eqn:x0}
\end{equation}
where $x_t^i$ represents the noisy latent feature, and $\epsilon_\theta(x_t^i)$ represents the estimated noise from the pre-trained diffusion model. At each denoising step $t$, we calculate $\hat{x}_0^i(x_t^i)$, representing the predicted $x_0^i$ and dubbed as the \textit{denoised observation} of $x_t^i$. $\alpha_t$ is the total noise variance parameterized via denoising step $t$.
\begin{figure}[t]
\centering
\includegraphics[width=\textwidth]{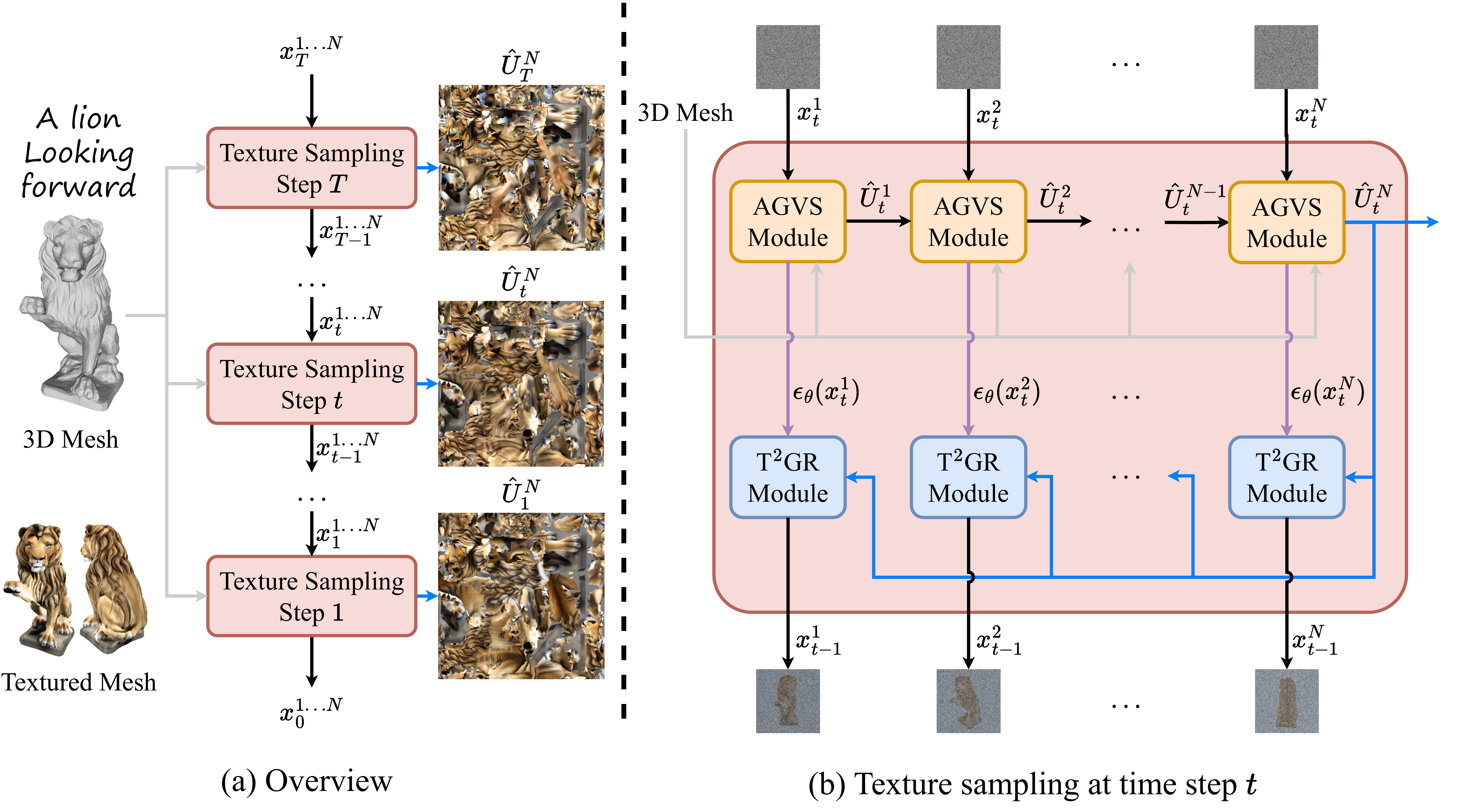}
\caption{Overview of our proposed method, where \textbf{AGVS} and \textbf{T$\mathbf{^2}$GR} denote Attention-Guided View Sampling and Text\&Texture-Guided Resampling, respectively. First of all, we sample $N$ viewpoints across the objects. As shown in (a), our texture sampling strategy is an interleaved process of texture generation and diffusion denoising. Specifically, our texture sampling process is structured into $T$ desnoising steps of diffusion process, and a complete RGB texture map ($\hat{U}_{t}^N$) is generated at the end of each step. As shown in (b), at denoising step $t$, each AGVS module receives noisy latent features $x_{t}^i$ as input to sample an image and produce a \textbf{partial texture map} $\hat{U}_{t}^i$, along with noise estimation $\epsilon_\theta(x_t^i)$. The generated $\hat{U}_{t}^i$ serves as guidance for sampling the subsequent view. Subsequently, a \textbf{complete texture map} $\hat{U}_{t}^N$ is employed to refine the noise estimation of each view within T$^2$GR modules, facilitating the prediction of noisy features for the ensuing denoising step ($x_{t-1}^{1...N}$).}
\label{fig:overview}
\end{figure}


\begin{figure}[t]
\centering
\includegraphics[width=0.68\textwidth]{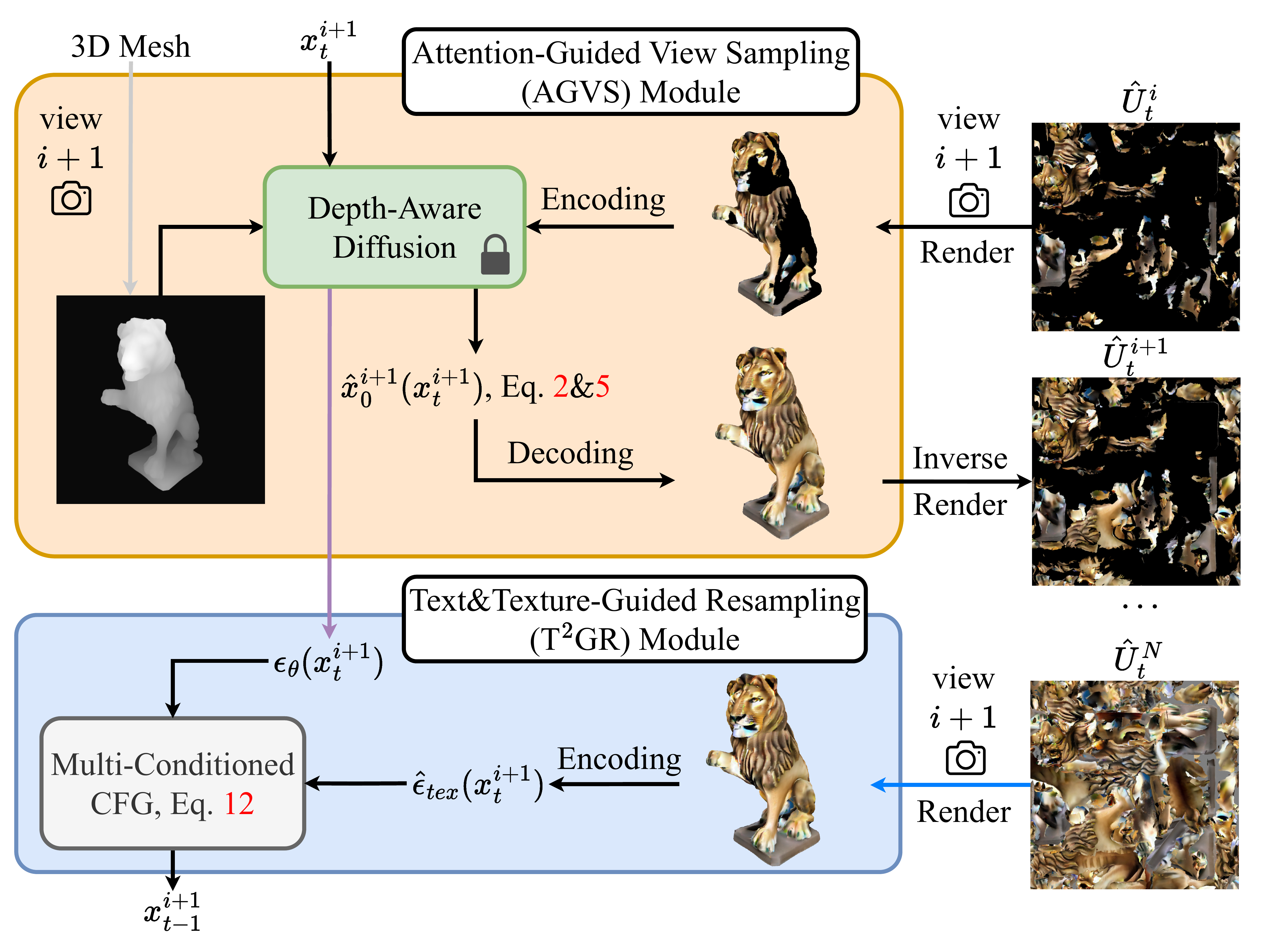}
\caption{Details of denoising for view $i+1$ at step $t$. The AGVS module is designed to generate denoised observation $\hat{x}_0^{i+1}(x_t^{i+1})$ which will be assembled onto UV space to form intermediate texture $\hat{U}_{t}^{i+1}$. The attention guidance is omitted in the figure for simplification. After iterating over all sampled views starting from $i=1$ to $N$, we obtain a complete texture map $\hat{U}_{t}^N$ for each denoising step. Conditioned on the current aggragated texture map, the T$^2$GR module will update the noise estimation $\epsilon_\theta(x_t^i)$ with the multi-conditioned classifier-free guidance (CFG) to calculate the noisy latent feature $x_{t-1}^{i+1}$ of the next denoising step.}
\label{fig:details}
\end{figure}

\noindent\textbf{Proposed Framework.} An overview of our proposed texture sampling is shown in Fig.~\ref{fig:overview}(a). It leverages the sequential nature of the denoising process of the diffusion model and maintains the 3D consistency of the generated texture at each denoising step. We adopt the similar strategy of interlaced multi-view sampling as in \cite{cao2023texfusion}. But instead of relying on a post-processing to convert sampled view consistent latents into RGB textures which resulted in over-smoothed textures, we directly enforce view consistent sampling in RGB texture space and develop an noise resampling strategy to retain rich texture details.

In detail, as shown in Fig.~\ref{fig:overview}(b) and Fig.~\ref{fig:details}, we execute the following two steps at each denoising step $t$.
First, by progressively assembling the \textit{denoised observations} $\hat{x}_0^i(x_t^i)$ where $i=1,\dots,N$, employing an attention-guided multi-view sampling strategy (Sec.~\ref{sec:attn}), we compute a view-consistent \textbf{noise-free} texture map $\hat{U}(x_t^{1\dots N})$.
For brevity, we denote $\hat{U}_{t}^i$ as the partial texture map $\hat{U}(x_t^{1\dots i})$ where $i<N$, and 
$\hat{U}_{t}^N$ as the complete texture map $\hat{U}(x_t^{1\dots N})$, both at denoising step $t$. Second, to retain rich texture details, we conduct a Text\&Texture-Guided Resampling step to calculate the noisy latent feature for the next denoising step $t-1$, conditioned on the current texture map $\hat{U}_{t}^N$ as well as the text-guided noise estimation from a pre-trained T2I model as elaborated in Sec.~\ref{sec:t2gr}.

Following the DDIM sampling, we go through the above process with $T$ denoising steps to arrive at the final generated texture map $\hat{U}_{1}^N$. We present the above-mentioned two major steps in the following sections.

\vspace{-1pt}
\subsection{Attention-Guided Multi-View Sampling}
\label{sec:attn}

As highlighted in Sec.~\ref{sec:intro}, conducting a full denoising process in sequence for each view generation, conditioned on previously observed views, can result in noticeable seams due to limited guidance from previous views. To mitigate this issue, we generate an RGB texture map at each denoising step with the \textit{denoised observation}. Since each denoising step of the diffusion model is conditioned on a complete texture map from the preceding denoising step, this significantly reduces view inconsistency.

In particular, we follow DDIM sampling, for each sampled view $i$ at denoising step $t$, the \textit{denoised observation} $\hat{x}_0^i(x_t^i)$ in the latent space can be computed as in Eq.~\ref{eqn:x0}. The \textit{denoised observation} $\hat{x}_0^i(x_t^i)$ are then decoded into images $I_t^i$ in the RGB space via the VAE decoder $\mathcal{D}$ of the pre-trained stable diffusion~\cite{rombach2022high}, 
\begin{equation}
I_t^i = \mathcal{D} (\hat{x}_0^i(x_t^i)).
\label{eqn:decode}
\end{equation}
Starting with the first viewpoint $i = 1$, we inverse render $I_t^i$ into the UV texture space, obtaining the partial texture map $\hat{U}_{t}^i$. Then for the subsequent viewpoint $i+1$, the prediction of its \textit{denoised observation} will depend on the current partial texture map.
More specifically, we render the partial texture map $\hat{U}_{t}^i$ onto viewpoint $i+1$, which is fed as input to the VAE encoder $\mathcal{E}$ to obtain the latent features $G_{t}^{i+1}$:
\begin{equation}
G_{t}^{i+1} = \mathcal{E}(Render^{i+1}(\hat{U}_{t}^{i})).
\label{eqn:inpainting_1}
\end{equation}
Referring to $\mathcal{M}^{i+1}$ as the mask delineating regions observed for the first time at view $i+1$ in RGB space, we adopt the approach of blended latent diffusion~\cite{avrahami2023blended} to fuse the encoded render $G_{t}^{i+1}$ with the original noisy latents $x_t^{i+1}$ using $\mathcal{M}^{i+1}$, which aims to solely generate unobserved regions while preserving observed ones. In particular, we align the \textbf{noise-free} render $G_{t}^{i+1}$ to the same noise level as $x_t^{i+1}$ by adding randomly sampled noise $\epsilon$ before blending. This process can be expressed as follows:
\begin{equation}
\begin{aligned}
x_t^{i+1} \leftarrow x_t^{i+1}\odot \mathcal{M}^{i+1}_\downarrow + (\sqrt{\alpha_t}\cdot G_{t}^{i+1} + \sqrt{1 - \alpha_t}\cdot\epsilon)\odot(1 - \mathcal{M}^{i+1}_\downarrow),
\end{aligned}
\label{eqn:inpainting_2}
\end{equation}
where $\downarrow$ symbolizes downsampling to the resolution of latent features. The revised $x_t^{i+1}$ is subsequently employed to compute the \textit{denoised observation} $\hat{x}_0^{i+1}(x_t^{i+1})$ for viewpoint $i+1$ at step $t$, in accordance with Eq.~\ref{eqn:x0}.

\begin{figure*}[t]
\centering
\includegraphics[width=0.97\textwidth]{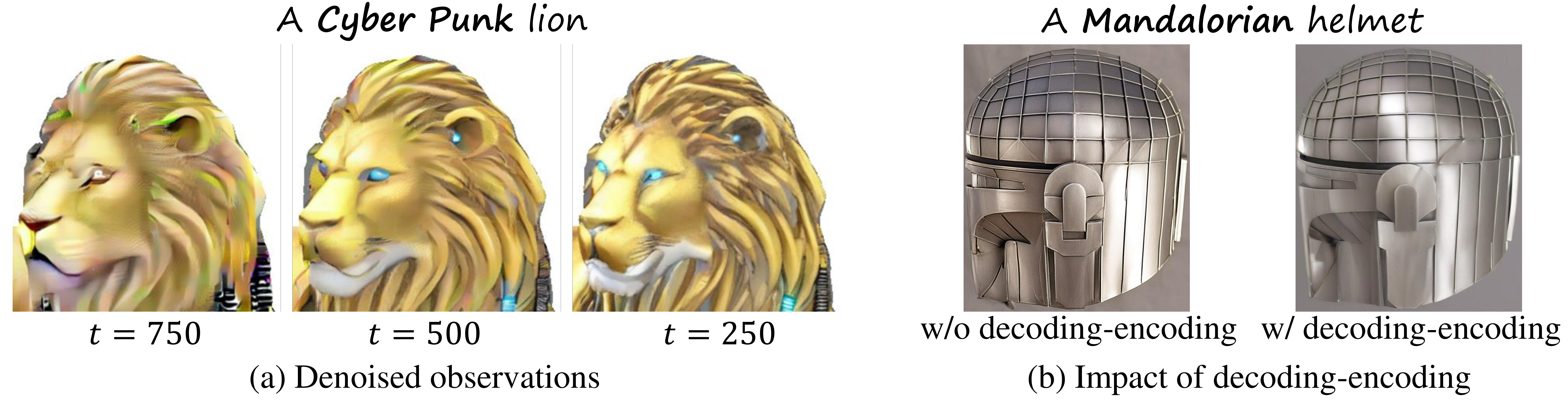}
\caption{(a) \textit{Denoised observation} $\hat{x_0}(x_t^i)$ of different denoising steps. The high-frequency information is gradually generated during sampling. (b) We claimed that the over-smoothness of directly using Eq.~\ref{eqn:correct} for noise sampling is caused by repeatedly going through VAE decoder and encoder at each denoising step. For validation, we conducted an ablation for a simplified case of generating only a single viewpoint. It shows that the over-smoothness still existed even for single view generation. Mathematically, if we do not have encoding and decoding operation at each denoising step, single view sampling is exactly same as DDIM sampling.}
\label{fig:inter}
\end{figure*}

Furthermore, as indicated in Fig.~\ref{fig:ab}(a), sequential generation across different viewpoints often falls short of ensuring consistent appearances, albeit without conspicuous seams. To address this, we introduce a novel attention-guided cross-view generation strategy. Drawing inspiration from the work of Cao~\etal~\cite{cao2023masactrl}, we believe the Key and Value features in the self-attention module of the stable diffusion encapsulate the local contents and textures of generated images. In detail, we regard the front view as the reference view and propagate the Key and Value of the reference view to other views. The process can be outlined as follows:
\begin{equation}
\epsilon_\theta(x_t^{ref}),\, Q_t^{ref}, \, K_t^{ref},\, V_t^{ref} \leftarrow Unet_\theta(x_t^{ref}),
\label{eqn:inpainting_3}
\end{equation}
\begin{equation}
\epsilon_\theta(x_t^i) \leftarrow Unet_\theta(x_t^i, K_t^{ref},V_t^{ref}).
\label{eqn:inpainting_4}
\end{equation}
Herein, $Q_t^{ref}$, $K_t^{ref}$, and $V_t^{ref}$ denote the Query, Key, and Value features from the self-attention module of the reference view, respectively.
In Eq.~\ref{eqn:inpainting_4}, the Key and Value features for each viewpoint are substituted with those from the reference view to calculate its estimated noise $\epsilon_\theta(x_t^i)$. Following this substitution, for each viewpoint 
$i$, the \textit{denoised observation} $\hat{x}_0^i(x_t^i)$ is updated in accordance with Eq.~\ref{eqn:x0}. 
As shown in Fig.~\ref{fig:inter}(a), the texture details will gradually appear in the \textit{denoised observation} as the diffusion process proceeds.

By sequentially applying the Eq.~\ref{eqn:decode}$\sim$Eq.~\ref{eqn:inpainting_2} on all viewpoints with our proposed attention-guided cross-view generation, we obtain a complete, view-consistent and noise-free texture map $\hat{U}_{t}^N$ for the current denoising step $t$.

\subsection{Text\&Texture-Guided Resampling (T$^2$GR)}
\label{sec:t2gr}
Upon obtaining the current texture map at step $t$, in this section we will present our Text\&Texture-Guided Resampling (T$^2$GR) approach for noise estimation to update the noisy latent features for the next denoising step $t-1$. 

As shown in Eq.~\ref{eqn:ddim_1}, the derivation of denoised latents $x_{t-1}^i$ depends on the estimated noise $\epsilon_{\theta}(x_{t}^{i})$ and the \textit{denoised observation} $\hat{x}_0^i(x_{t}^{i})$. Given that $\hat{x}_0^i(x_{t}^{i})$ is expected to exhibit view consistency maintained by the texture map $\hat{U}_{t}^N$, recalculating the noise map $\epsilon_{\theta}(x_{t}^{i})$ under the guidance of $\hat{U}_{t}^N$ ensures to preserve the view consistency. Specifically, in Eq.~\ref{eqn:x0} we set $\hat{x}_0^i(x_{t}^{i})$ equal to the current encoded render of the texture map $\hat{U}_{t}^N$ at view $i$. From this, we derive the recalculated noise map $\hat{\epsilon}_{tex}(x_t^i)$ as follows:

\begin{equation}
\hat{\epsilon}_{tex}(x_t^i) = \frac{x_t^i - \sqrt{\alpha_t}\cdot\mathcal{E}(Render^i(\hat{U}_{t}^N))}{\sqrt{1 - \alpha_t}}.
\label{eqn:correct}
\end{equation}
This recalculated noise map is then utilized in place of $\epsilon_{\theta}(x_{t}^{i})$ in Eq.~\ref{eqn:ddim_1} and Eq.~\ref{eqn:x0} for the computation of $x_{t-1}^i$.

While the above noise map update strategy ensures view consistency, it tends to result in over-smoothed images (as shown in Fig.~\ref{fig:inter}(b) and Fig.~\ref{fig:ab}(b)). This is primarily because the VAE encoder $\mathcal{E}$ in the stable diffusion model compresses high-frequency details, referred to as \textit{imperceptible details}, as noted by ~\cite{rombach2022high}. The repeated use of the encoder $\mathcal{E}$ leads to an accumulation of this detail compression, affecting the overall image quality.

To avoid over-smoothness, we utilize the text-guided noise estimation which is not affected by the repeatedly encoding and decoding operation. Meanwhile, we take the current texture map as an additional condition to derive a multi-conditioned noise estimation formulation. The text-guided noise $\epsilon_\theta(x_t^i|c)$ can be directly computed from the diffusion model and now we need to compute the texture-conditioned noise estimation which we denote as $\epsilon_{tex}(x_t^i|\hat{U}_{t}^N)$. By analyzing the formulation of $\epsilon_\theta(x_t^i)$, we see that it is essentially a weighted combination of conditional noise prediction $\epsilon_\theta(x_t^i|c)$ and unconditional noise prediction $\epsilon_\theta(x_t^i|\varnothing)$, following the Classifier-Free Guidance (CFG) introduced in~\cite{ho2022classifier}:
\begin{equation}
\epsilon_\theta(x_t^i) = \epsilon_\theta(x_t^i|\varnothing) + \omega (\epsilon_\theta(x_t^i|c) - \epsilon_\theta(x_t^i|\varnothing)),
\label{eqn:cfg_c}
\end{equation}
where $c$ and $\varnothing$ represent the text prompt and null-text prompt, respectively, and $\omega$ is a user-specified weight. 
Similarly, the $\epsilon_{tex}(x_t^i)$ is assumed to follow the same formulation of CFG, 
\begin{equation}
\epsilon_{tex}(x_t^i) = \epsilon_\theta(x_t^i|\varnothing) + \omega (\epsilon_{tex}(x_t^i|\hat{U}_{t}^N) - \epsilon_\theta(x_t^i|\varnothing)).
\label{eqn:cfg_uv}
\end{equation}

Thus, to disentangle the texture-conditioned noise estimation $\epsilon_{tex}(x_t^i|\hat{U}_{t}^N)$, we subtract the null-text conditioned noise estimation from $\epsilon_{tex}(x_t^i)$. Here we have $\epsilon_{tex}(x_t^i) = \hat{\epsilon}_{tex}(x_t^i)$ from Eq.\ref{eqn:correct}. The computation for the texture-conditioned noise estimation $\epsilon_{tex}(x_t^i|\hat{U}_{t}^N)$ is as follows:
\begin{equation}
\epsilon_{tex}(x_t^i|\hat{U}_{t}^N) = \frac{1}{\omega}(\hat{\epsilon}_{tex}(x_t^i)  - \epsilon_\theta(x_t^i|\varnothing)) + \epsilon_\theta(x_t^i|\varnothing).
\label{eqn:uv_cond}
\end{equation}

In the end, we formulate our multi-conditioned CFG for final noise estimation, which is conditioned on both the textual prompt and texture map:
\begin{equation}
\begin{aligned}
\epsilon_m(x_t^i) = \epsilon_\theta(x_t^i|\varnothing)+ \omega_1 (\epsilon_\theta(x_t^i|c) - \epsilon_\theta(x_t^i|\varnothing)) +\omega_2 (\epsilon_{tex}(x_t^i|\hat{U}_{t}^N) - \epsilon_\theta(x_t^i|\varnothing)),
\end{aligned}
\label{eqn:multi_cond}
\end{equation}
where $\omega_1 + \omega_2 = \omega$. We exploit a large $\omega_2$ for early sampling steps, which will decrease linearly from $\omega$ to $0$ in the process of denoising. The comprehensive derivation of Eq.~\ref{eqn:multi_cond} can be found in the supplementary materials.
Finally, we compute $x_{t-1}^i$ for the subsequent denoising step by letting $\epsilon_\theta(x_t^i) = \epsilon_m(x_t^i)$ in Eq.~\ref{eqn:ddim_1} and Eq.~\ref{eqn:x0}. 

By combining the attention-guided multi-view sampling and text\&texture-guided resampling, our proposed method can directly generate high-fidelity and view consistent texture map in the RGB space, without the need of an MLP post-processing as in TexFusion~\cite{cao2023texfusion} which results in over-smoothed textures.


\begin{figure*}[t]
\centering
\includegraphics[width=0.98\textwidth, page=1]{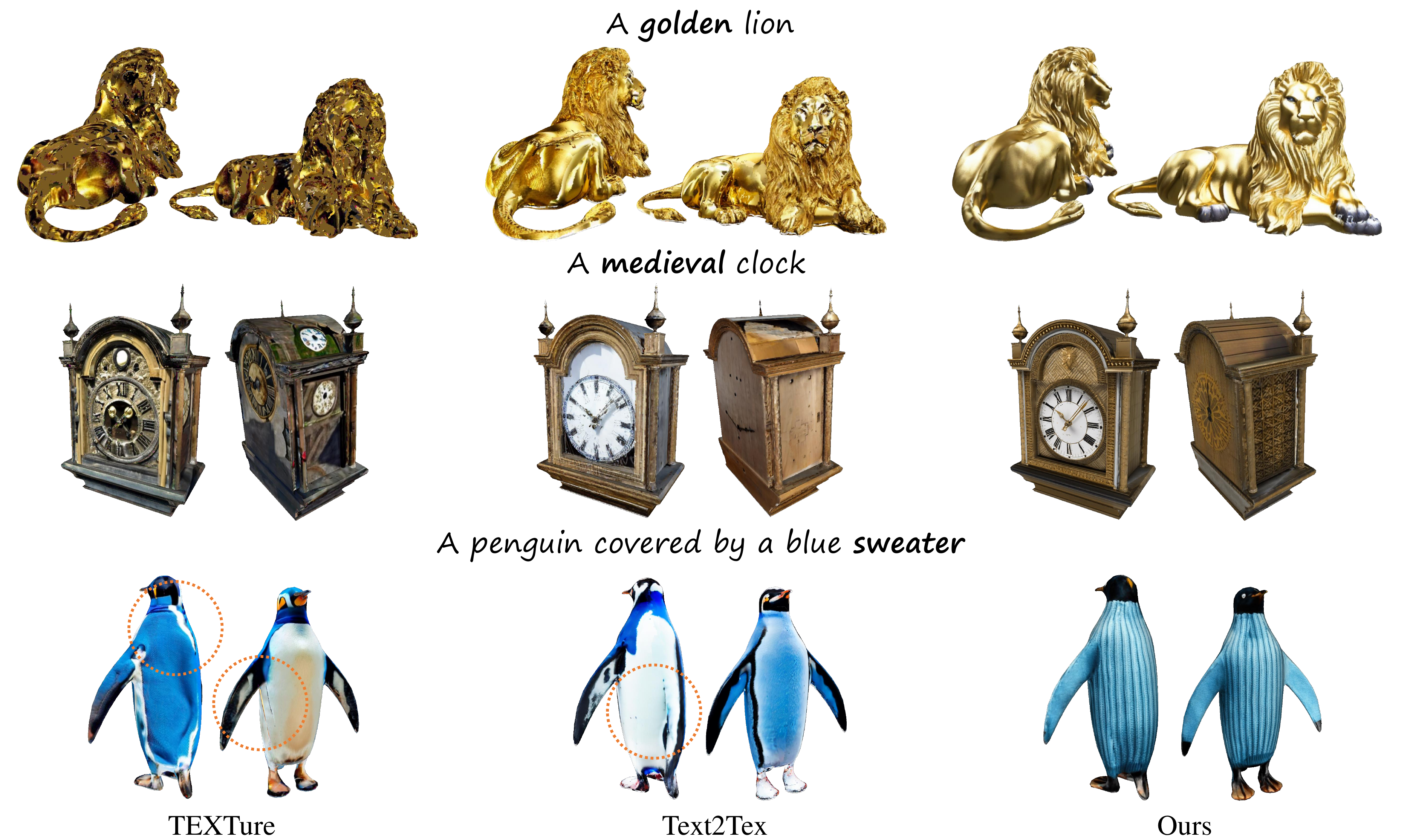}

\caption{Visual comparison of our proposed method against TEXTure~\cite{richardson2023texture} and Text2Tex~\cite{chen2023text2tex}.}
\label{fig:qua_1}
\end{figure*}
\begin{figure*}[t]
\centering
\includegraphics[width=0.98\textwidth]{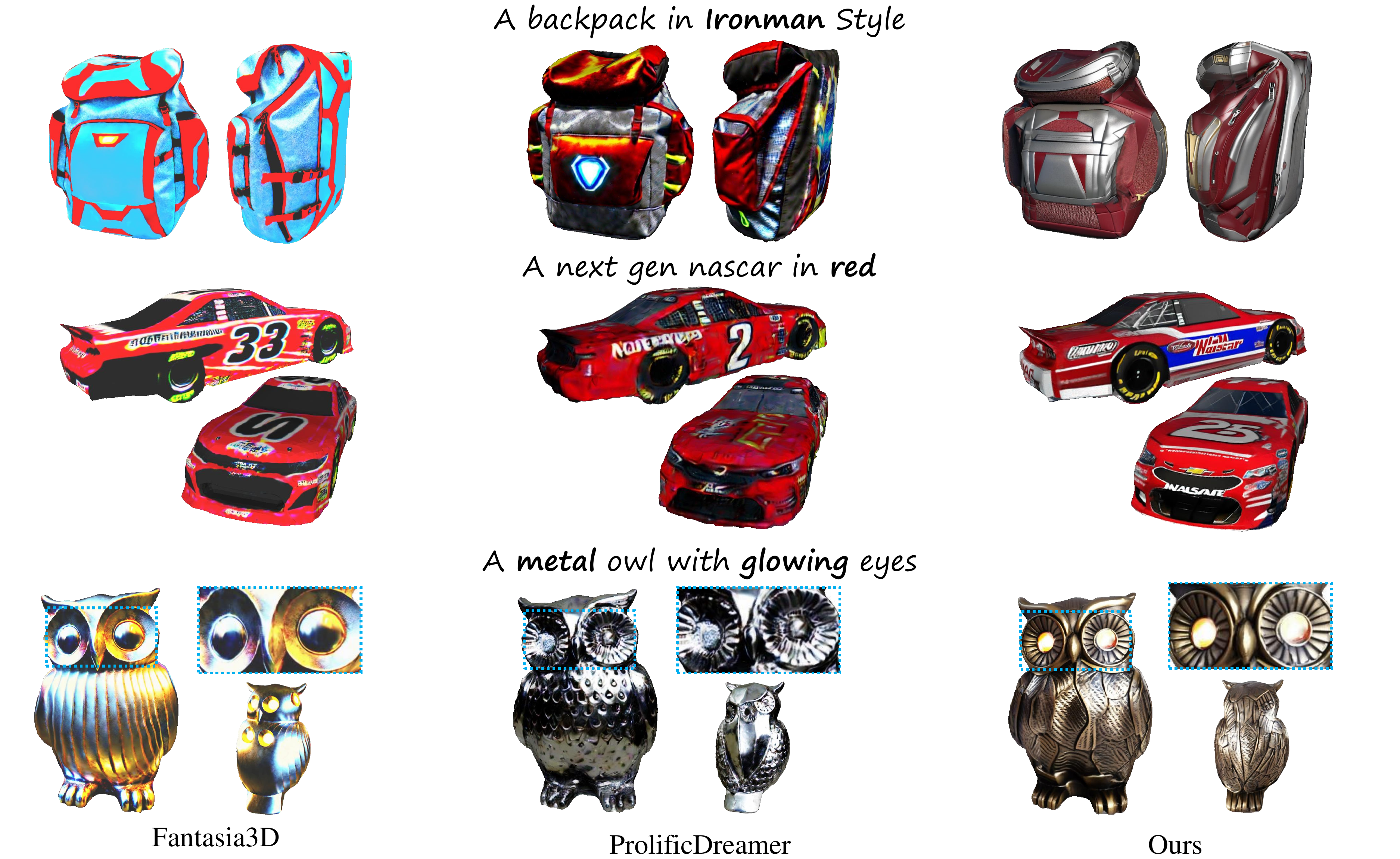}
\caption{Visual comparison of our proposed method against Fantasia3D~\cite{chen2023fantasia3d} and ProlificDreamer~\cite{wang2023prolificdreamer}.}
\label{fig:qua_2}
\end{figure*}
\begin{figure}[!htbp]
\centering
\includegraphics[width=0.98\textwidth]{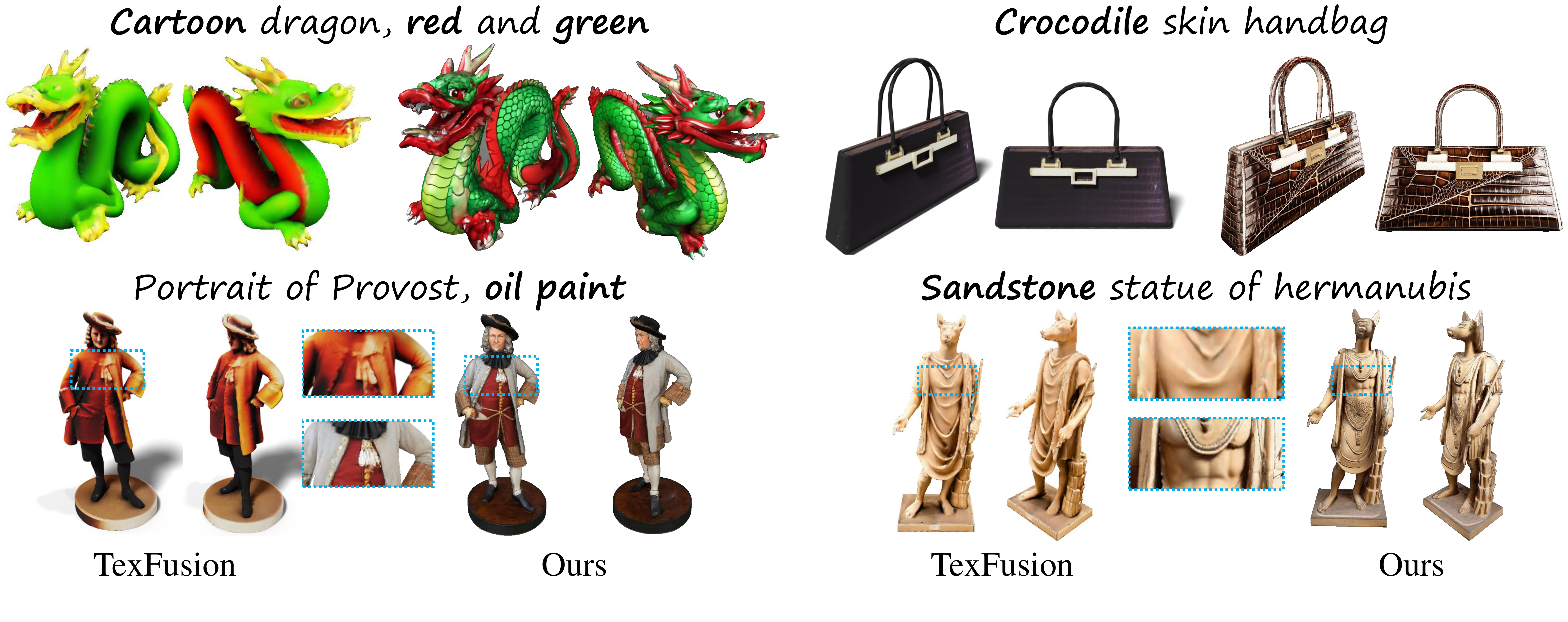}
\caption{Visual comparison of our proposed method against TexFusion~\cite{cao2023texfusion}. The results of TexFusion are directly copied from its original paper.}
\label{fig:qua_3}
\end{figure}

\section{Experiments}
\label{sec:exp}
\subsection{Implementation Details}

We employ the depth-aware diffusion model provided by ControlNet~\cite{zhang2023adding} as our T2I backbone with denoising steps $T=40$. To render objects, we take eight different viewpoints around the object. The pose is sampled in spherical coordinates, with elevation angles being zero and azimuth angles uniformly sampled between [0$^{\circ}$, 360$^{\circ}$]. An additional top view is sampled.
Additionally, we employ the Xatlas~\cite{xatlas2016} tool to compute the UV atlas for a given mesh. 

\noindent\textbf{Dataset.} Our experiments incorporate a diverse collection of 45 meshes, sourced from various datasets such as Objaverse~\cite{deitke2023objaverse} and ThreeDScans~\cite{threedscans}, with 2 to 3 distinct prompts for each mesh. Please refer to the supplementary for details.

\subsection{Compared Methods} 
We conduct experimental comparison over several state-of-the-art approaches, including TEXTure~\cite{richardson2023texture}, Text2Tex~\cite{chen2023text2tex}, Fantasia3D~\cite{chen2023fantasia3d},  ProlificDreamer~\cite{wang2023prolificdreamer} and TexFusion~\cite{cao2023texfusion}. For TEXTure, Text2Tex, and Fantasia3D, we use their respective publicly available codebase. For ProlificDreamer, we adopt the implementation of ThreeStudio~\cite{threestudio2023} and replaced its backbone with ControlNet~\cite{zhang2023adding} to recognize the depth. In the case of TexFusion, where the implementation is not available, our analysis is limited to a qualitative assessment using results extracted directly from the original paper. Notably, for all the compared approaches, the geometry remains fixed during texture generation.

\subsection{Qualitative Comparison}
We provide visual comparison in Fig.~\ref{fig:qua_1} and Fig.~\ref{fig:qua_2}. 
Specifically, in Fig.~\ref{fig:qua_1}, we showcase the robustness of our approach in addressing fragmented textures against progressively texture assembling approaches, namely TEXTure~\cite{richardson2023texture} and Text2Tex~\cite{chen2023text2tex}. 
This improvement is credited to our use of attention-guided view sampling combined with a distinct text\&texture-guided resampling approach, which maintain view consistency at each denoising step to persistently enhance 3D consistency. 

\begin{table}[t]
\caption{Quantitative comparison on generated textures.}
\centering
\scriptsize
\setlength\tabcolsep{2.5pt}
\begin{tabular}{lccc}
\toprule
Methods                       & FID~$\downarrow$   & KID$\times10^{-3}\downarrow$  & CLIPScore~$\uparrow$  \\ 
\midrule
TEXTure                  & 99.06 & 7.23 & 19.73 \\
Text2Tex                  & 109.94 & 7.17 & 21.26\\ 
Fantasia3D                  & 108.58 & 7.52 & 21.14 \\ 
ProlificDreamer                  & 94.51 & 7.00 & 21.25 \\ 
\midrule
Ours                   & \textbf{84.65} & \textbf{4.27} & \textbf{22.83} \\
\bottomrule
\end{tabular}
\label{tab:quan}
\end{table}
\begin{table}[!htbp]
\caption{User Study Preference: The entries in the table indicate our preference over other methods. A higher value represents a greater preference.}
\centering
\scriptsize
\setlength\tabcolsep{2.5pt}
\begin{tabular}{lcccc}
\toprule
                           &  TEXTure~$\uparrow$ & Text2Tex~$\uparrow$ & Fantasia3D~$\uparrow$ & ProlificDreamer~$\uparrow$     \\ 
\midrule
Ours            & 64.72\% & 71.46\% & 70.97\% & 69.18\% \\ 
\bottomrule
\end{tabular}
\label{tab:user_study}
\end{table}

In Fig.~\ref{fig:qua_2}, we compare with score distillation based approaches, namely Fantasia3D~\cite{chen2023fantasia3d} and ProlificDreamer~\cite{wang2023prolificdreamer}. As demonstrated, Fantasia3D typically produces textures that are over-smoothed and over-saturated, while ProlificDreamer, though more detailed and contrasted, is marred by evident artifacts of blurry edges. In contrast, our method surpasses these distillation-based methods by generating more realistic high-quality results.

\noindent\textbf{Comparison with TexFusion~\cite{cao2023texfusion}.} 
We also present a qualitative comparison of our method with TexFusion~\cite{cao2023texfusion} in Fig.~\ref{fig:qua_3}. TexFusion employed instantNGP~\cite{muller2022instant} to mitigate inconsistencies post-decoding of latent features into RGB space, which often led to over-smoothed results. In contrast, our method effectively generates textures that are consistent across views and retain rich details. Please refer to supplementary materials for more visual results.

\subsection{Quantitative Comparison}
\noindent\textbf{Evaluation Metrics.} For quantitative evaluation of the generated texture, we employ two widely used image quality and diversity evaluation metrics, including Frechet Inception Distance (FID)~\cite{heusel2017gans} and Kernel Inception Distance (KID)~\cite{binkowski2018demystifying}. 
These metrics are instrumental in measuring the distribution similarity between two sets of images. For each comparison method, we render a set of images by uniformly sampling 32 different views of the generated textured mesh. 
To establish a ground truth image set, we follow the approach outlined by Cao~\etal~\cite{cao2023texfusion} which used a depth-conditioned ControlNet to synthesize images conditioned on rendered depth maps and corresponding textual prompts. 
The background pixels have been removed from all images to mitigate the influence caused by unconstrained background.
Additionally, we incorporate the CLIPScore metric~\cite{hessel2021clipscore} to assess the congruence and resemblance between the generated images and their associated text prompts. Specifically, for each method, we calculate the average CLIPScore across all rendered images relative to the given text prompts.

We present the quantitative evaluations of the above-mentioned methods on FID, KID and CLIPScore in Tab.~\ref{tab:quan}. Notably, our approach demonstrates superior performance, outstripping the other methods by at least $10.4\%$ in FID and $39.0\%$ in KID. The figures showcase our method's capability to generate textures that not only are more realistic but also exhibit a wide variety of appearances across diverse objects.

\noindent\textbf{User Study.} 
To analyze the quality of the generated textures and their fidelity to the corresponding text prompts, we conducted a detailed user study of our method against four baseline methods. We randomly select 40 meshes from our collected data and feed them along with a text prompt as the input for each method. For each of these 40 selections, we generate 360$^{\circ}$ rotating view videos using both our method and one of the baseline methods and display them side-by-side. Participants in the study are then requested to select the video that not only better matched the given caption but also exhibited superior quality. The user study yielded a dataset of 2,480 responses from 62 participants. We report the user preferences in Tab.~\ref{tab:user_study}. The results indicate that our method is notably more effective in producing high-quality textures that are preferred by human evaluators.

\begin{table}[t]
\caption{Ablation study over attention guidance from attention-guided cross-view generation and T$^2$GR. }
\centering
\scriptsize
\setlength\tabcolsep{2.5pt}
\begin{tabular}{lcc}
\toprule
Methods                       & FID~$\downarrow$   & KID$\times10^{-3}\downarrow$ \\ 
\midrule
Ours w/o attention guidance                  & 98.62 & 5.01 \\
Ours w/ $(\omega_1 = 0)$      & 95.97 & 4.60 \\ 
Ours w/ $(\omega_2 = 0)$      & 99.46 & 5.22 \\ 
\midrule
Ours                   & \textbf{84.65} & \textbf{4.27} \\
\bottomrule
\end{tabular}
\label{tab:ab}
\end{table}
\begin{figure*}[t]
\centering
\includegraphics[width=0.95\textwidth, page=1]{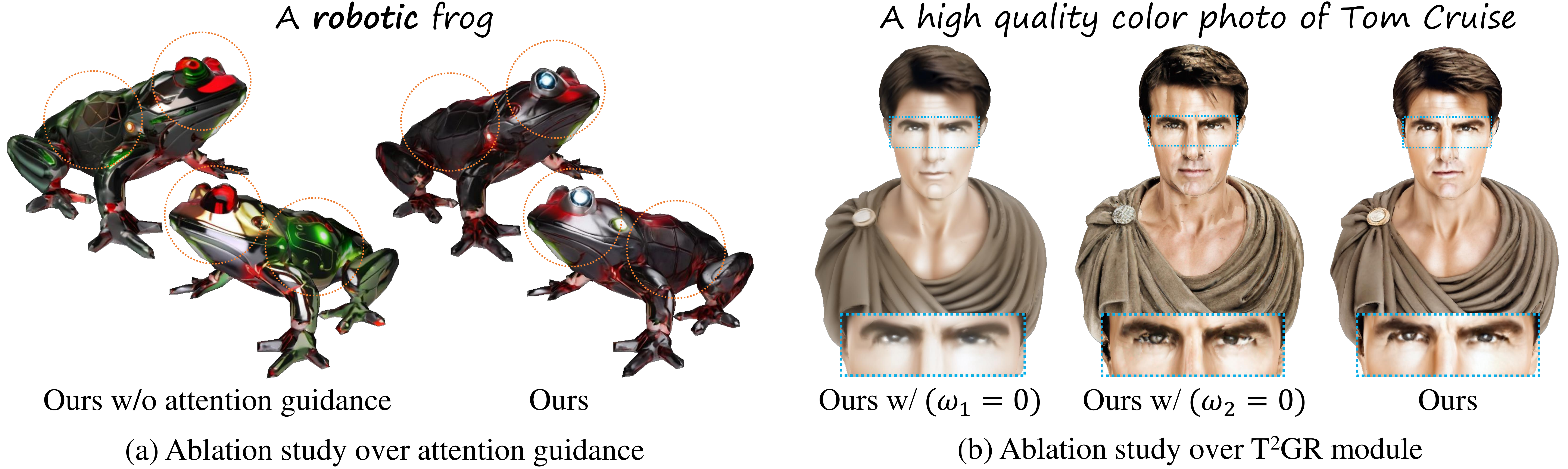}
\caption{Visual comparison of ablation study over (a) attention guidance from attention-guided cross-view generation and (b) T$^2$GR module. Without attention guidance, the frog has different appearance patterns and color tones over different sides. }
\label{fig:ab}
\end{figure*}

\subsection{Ablation Study} 

We first visually evaluate the impact of our proposed attention-guided cross-view generation as shown in Fig.~\ref{fig:ab}(a). The results demonstrate that our proposed method with attention guidance is able to generate textures which have a consistent appearance in different viewpoints. We also evaluate the impact of the T$^2$GR by keeping $\omega_1 = 0$ and $\omega_2 = 0$ in Eq.~\ref{eqn:multi_cond}, respectively. As shown in Fig.~\ref{fig:ab}(b), the first figure with $\omega_1 = 0$ lacks high-frequency details and tends to be over-smoothed, while the middle figure with $\omega_2 = 0$ lacks texture guidance and the assembled texture from all viewpoints is fragmented. We also evaluate the generation quality using FID and KID in Tab.~\ref{tab:ab}, which shows that our method with attention-guided cross-view
generation and T$^2$GR outperforms other variants by a large margin. More ablation studies about the impact of reference views are shown in the supplementary materials.

\begin{figure}[t]
\centering
\includegraphics[width=0.95\textwidth]{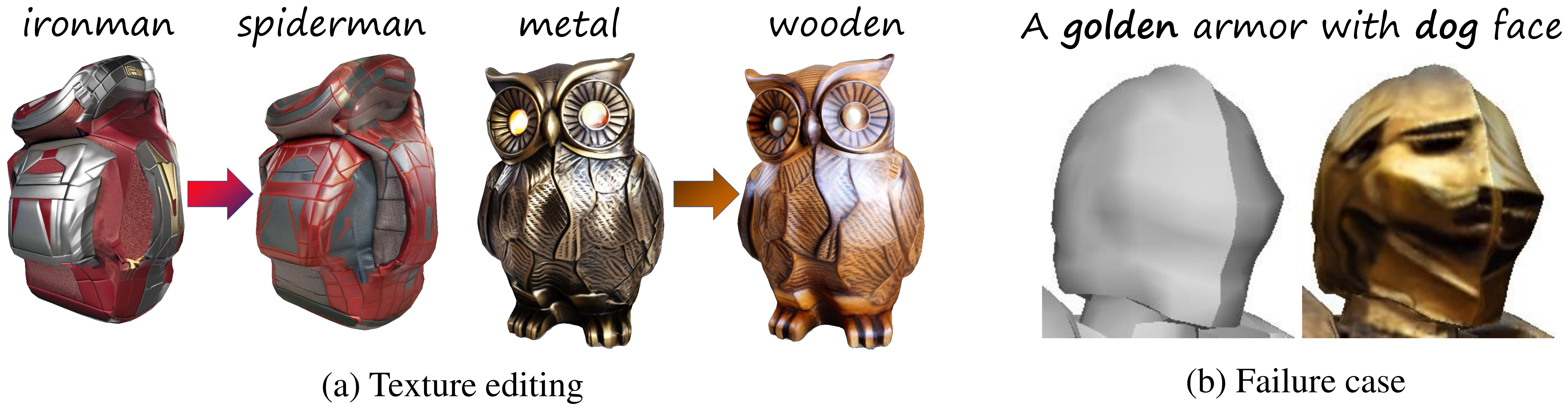}
\caption{(a) Applications of our proposed texture sampling strategy for text-driven texture editing. (b) Our method struggles to generate correct textures when text prompts are far from the semantics of the mesh geometry. }
\label{fig:edit}
\end{figure}

\subsection{Application and Failure Case}

Our proposed texture sampling scheme can also be applied to texture editing, as shown in Fig.~\ref{fig:edit}(a). It shares the same pipeline with texture generation, but here we replace the depth-aware ControlNet with the MultiControlNet~\cite{zhang2023adding} that combines both the depth-guided and edge-guided generation to preserve the original identity, where the Canny edges are extracted from the generated views.
Fig.~\ref{fig:edit}(b) show a failure case of our method caused by significant semantic mismatch between text prompts and mesh geometry.

\section{Conclusion}
\label{sec:conclusion}

In this paper, we present TexGen, a novel texture sampling strategy for text-driven texture generation on 3D meshes, leveraging depth-aware diffusion models.
To address the significant challenges in producing textures that are consistent across views and rich in detail, we first propose to maintain a time-dependent texture map that evolves with each denoising step to progressively reduce the view discrepancy. Specifically, at each denoising step, the texture is assembled from the \textit{denoised observations} of sampled views under our attention-guided multi-view sampling process. It is then utilized in our text\&texture-guided noise resampling procedure to further guide the estimated noise fed into the next denoising step.
The effectiveness of our method is evident in its ability to generate superior-quality textures for diverse 3D objects as well as in its adaptability for texture editing purposes.
As for limitations, the overall quality of the generated 3D textures still exhibits a gap when compared to 2D image generation. Striking a balance between 3D consistency and the generation quality of specific views remains a challenge. Additionally, we didn't consider the disentanglement of material and lighting from the generated textures, which is leaved as future work to explore.



\paragraph{{\rm {\bf Acknowledgement.}}} We gratefully acknowledge the support of MindSpore (https://www.mindspore.cn/), CANN (Compute Architecture for Neural Networks) and Ascend AI Processor used for this research. Huo and Yang would like to thank the financial support from the Natural Sciences and Engineering Research Council of Canada and the University of Alberta.

%
%
\bibliographystyle{splncs04}
\bibliography{main}
\end{document}